\documentclass[10pt, a4paper]{article}
\usepackage{lrec2022} 
\usepackage{multibib}
\newcites{languageresource}{Language Resources}
\usepackage{graphicx}
\usepackage{tabularx}
\usepackage{soul}
\usepackage{float}
\usepackage{caption}
\usepackage{booktabs}
\usepackage{multirow}
\usepackage{titlesec}
\titleformat{\section}{\normalfont\large\bfseries\center}{\thesection.}{1em}{}
\titleformat{\subsection}{\normalfont\SmallTitleFont\bfseries\raggedright}{\thesubsection.}{1em}{}
\titleformat{\subsubsection}{\normalfont\normalsize\bfseries\raggedright}{\thesubsubsection.}{1em}{}
\renewcommand\thesection{\arabic{section}}
\renewcommand\thesubsection{\thesection.\arabic{subsection}}
\renewcommand\thesubsubsection{\thesubsection.\arabic{subsubsection}}

\usepackage{epstopdf}
\usepackage[utf8]{inputenc}

\usepackage{hyperref}
\usepackage{xstring}

\usepackage{color}

\title{Documenting Geographically and Contextually Diverse Data Sources: The BigScience Catalogue of Language Data and Resources}

\name{Angelina McMillan-Major$^{4,7}$, Zaid Alyafeai$^{10}$, Stella Biderman$^{1,3}$, Kimbo Chen, \\
\large\textbf{Francesco De Toni$^8$, G\'{e}rard Dupont, Hady Elsahar, Chris Emezue, Alham Fikri Aji,}\\
\large\textbf{Suzana Ili\'{c}$^4$, Nurulaqilla Khamis, Colin Leong, Maraim Masoud, Aitor Soroa$^9$,}\\
\large\textbf{Pedro Ortiz Suarez$^{5,6}$, Zeerak Talat, Daniel van Strien$^2$, Yacine Jernite$^4$}} 

\address{Booz Allen Hamilton$^1$, British Library$^2$, EleutherAI$^3$, Hugging Face$^4$, Inria$^5$, Sorbonne Université$^6$, \\ University of Washington$^7$, University of Western Australia$^8$ ,
         University of the Basque Country$^9$,
         KFUPM $^{10}$\\
         angie@huggingface.co,
         francesco.detoni@uwa.edu.au, 
         pedro.ortiz@inria.fr,
         daniel.van-strien@bl.uk\\}

\abstract{
In recent years, large-scale data collection efforts have prioritized the amount of data collected in order to improve the modeling capabilities of large language models. 
This prioritization, however, has resulted in concerns with respect to the rights of data subjects represented in data collections, particularly when considering the difficulty in interrogating these collections due to insufficient documentation and tools for analysis.
Mindful of these pitfalls, we present our methodology for a documentation-first, human-centered data collection project as part of the BigScience initiative. 
We identified a geographically diverse set of target language groups (Arabic, Basque, Chinese, Catalan, English, French, Indic languages, Indonesian, Niger-Congo languages, Portuguese, Spanish, and Vietnamese, as well as programming languages) for which to collect metadata on potential data sources.
To structure this effort, we developed our online catalogue as a supporting tool for gathering metadata through organized public hackathons. 
We present our development process; analyses of the resulting resource metadata, including distributions over languages, regions, and resource types; and our lessons learned in this endeavor. 
 \\ \newline  }

\begin{document}

\maketitleabstract

\section{Introduction}

Current trends in developing large-scale language models require the use of vast amounts of data \cite{brown2020language,gao2020pile,rae2021scaling}. Typically, this data is collected from online sources, ranging from highly edited and structured text such as Wikipedia to the myriad text and audiovisual components of web pages collected by Common Crawl\footnote{\url{http://commoncrawl.org/}}. 
Several issues concerning their creation and the implications of their use, however, have been raised in recent research.
For instance, Wikipedia is highly biased in terms of the topics covered and in terms of the demographics of its contributors, particularly for gender, race, and geography \cite{barera2020mind}, resulting in similar concerns of representation in technologies developed on Wikipedia data.
Common Crawl, meanwhile, has been shown to contain hate speech and over-represent sexually explicit content \cite{luccioni-viviano-2021-whats}, and typical web-crawling collection practices have no structures for supporting informed consent beyond websites' own terms and conditions policies that users rarely read \cite{Cakebread,obar2020tos}.

Several documentation schemas for natural language processing (NLP) datasets \cite{bender-friedman-2018-data,gebru2018datasheets,gebru2021datasheets,holland2018dataset,pushkarna2021datacardsplaybook} have been recently produced to aid NLP researchers in documenting their own datasets \cite{gao2020pile,pile-datasheet,gehrmann2021gem,wang2021adversarial} and even to retrospectively document and analyze datasets that were developed and released by others without thorough documentation \cite{bandy2021addressing,kreutzer2021quality,birhane2021multimodal,dodge-etal-2021-documenting}. Data documentation to support transparency has gained traction following calls for a reevaluation of the treatment of data in machine learning (ML) at large \cite{prabhu2020large,jo_gebru2020archives,paullada2021data,gebru2021datasheets}. Building on this work, we focus on notions of representation, consent, transparency, self-determination, and privacy in a documentation-first, human-centered approach to data collection for NLP.
In this way, we aim to create a dataset for large multilingual language models that is responsibly collected and emphasises data subjects' rights to control over their own data.

\subsection{The BigScience Research Workshop}

Our work is situated within the BigScience workshop\footnote{\url{https://bigscience.huggingface.co/}}, a large scale coalition of experts in NLP and related fields from around the world. While the workshop has many working groups with different focuses, one of the primary goals of the project as a whole is to train and publicly release a large multilingual language model. A key part of accomplishing this became the creation of a dataset to train the model on.

With the limitations of previous large-scale data collection methods in mind, we set out to intentionally curate the BigScience dataset for \textit{representativeness}.
For our purposes, we define representativeness based on the intersection of geographic and sociolinguistic context. This means that, for each target language, we aim to collect data for the relevant dialects and regions where that language is spoken.
Starting from the goal of geographic representativeness and the BigScience members' languages of expertise, we identified 13 language groups to target for inclusion in the model training, namely Arabic, Basque, Chinese, Catalan, English, French, Indic languages, Indonesian, Niger-Congo languages, Portuguese, Spanish, and Vietnamese, as well as programming languages. 
Like most language modeling endeavors, we actively sought commonly used web sources for collection, but we also highlighted the need for other formats, including books, audio from radio programs and podcasts, and others.

To prepare for the challenges of responsible dataset creation, we focused our efforts on documenting potential sources prior to collection, while working groups in data governance and data tooling created plans for appropriate hosting and processing of the identified resources.
We compare this documentation effort, which we call the BigScience catalogue, to prior catalogs developed in linguistics and NLP (\S2). 
We present our online form\footnote{Available at \url{http://bigscience.huggingface.co/data-catalogue-form}} for the catalogue (\S3), developed to facilitate organized hackathon events for collecting metadata for sources from specific regions which we made open to public participants outside of BigScience (\S4). 
While we continue to accept submissions to the online form, we present the results of this initial effort in \S5.
Following the analysis of the results, we discuss challenges and limitations (\S6) and suggest improvements for future data documentation efforts (\S7). 

\section{Related Work}


Since the early 90s, NLP data organizations have maintained datasets and tools in order to support language research. 
Such organizations include the Linguistic Data Consortium (LDC) \footnote{\url{https://www.ldc.upenn.edu}}, the European Language Resources Association\footnote{\url{http://www.elra.info}}, the Chinese LDC\footnote{\url{http://www.chineseldc.org}}, the LDC for Indian Languages\footnote{\url{http://www.ldcil.org}}, and CLARIN\footnote{\url{https://www.clarin.eu}}. 
These organizations distribute licensed language resources such as annotated corpora and lexicons primarily to institutions, which pay to provide access to these datasets and supporting tools for their members.
The fees paid to the organizations support the creation, licensing, storage, and maintenance of new datasets and language research initiatives.
The LDC, for example, currently provides access to 841 datasets with 96 datasets added between 2018 and 2020 \cite{cieri-etal-2020-progress}.

Open source dataset catalogs have also been constructed as supporting technical infrastructure in the context of NLP and ML libraries. 
The Natural Language Toolkit (NLTK), developed since 2001, is a Python package with utilities for NLP tasks that includes access to widely used corpora such as the Brown Corpus \citelanguageresource{browncorpus} as well as features for adding datasets and using datasets locally \cite{NLTK}.
The Hugging Face Datasets library \cite{lhoest-etal-2021-datasets} and Tensorflow library \cite{TFDS} both provide tools for loading datasets from online as well as locally and include catalogs of directly accessible datasets.
The Southern African Centre for Digital Language Resources\footnote{\url{https://sadilar.org}} provides its own catalogue of annotated language datasets as well as processing tools, with links for downloading when resources are licensed for distribution.
Other catalogs of NLP datasets do not provide access to the datasets themselves, but provide information about uses and categories.
For example, Papers with Code links academic publications that use the same dataset with information about the dataset.\footnote{\url{https://paperswithcode.com/datasets}}
Masader, developed by BigScience members prior to the organization of the hackathons, similarly provides metadata about Arabic-language NLP datasets without hosting the data \cite{alyafeai2021masader}. 

\section{The Catalogue}
The main goal of the catalogue is to support the creation of the BigScience dataset while adhering to the values laid out by the various data working groups: collecting diverse resources (Data Sourcing), supporting information required for open and easily usable technical infrastructure (Data Tooling), and respecting the privacy of data subjects and the rights of data owners (Data Governance).
We collected the metadata topics for the dataset outlined by these working groups, resulting in almost 40 items, which we then grouped and prioritized.
While choosing the metadata items, we tried to balance the information needs of the working groups and the effort required on the part of the person submitting a resource to the catalogue while also prioritizing metadata that would be appropriate across languages and data sources.

In order to streamline the process for creating the catalogue, we decided to create an openly accessible online form for people to submit metadata for suggested resources.
We used an iterative design approach to collectively develop questions to elicit the metadata, descriptions explaining what information is being requested, and likely answers.
Wherever possible, we formatted the questions as multiple choice questions with the option for a free response if the appropriate answer was not available. 
After building the online form using Streamlit\footnote{\url{https://streamlit.io/}}, we tested the form with actual examples, such as \textit{Le Monde} newspaper and its publishing company, \textit{Group Le Monde}.

\subsection{The Catalogue Submission Form}
In testing the form with different resources, we realized that not all of the questions were necessary or appropriately worded for some kinds of resources, particularly those aimed at understanding how to process the data in the resource.  
With this consideration in mind, we defined the following resource types:
\textbf{primary source}, a single source of language data (text or speech), such as a newspaper, radio, website, or book collection;
\textbf{processed language dataset,} a processed NLP dataset containing language data that can be used for language modeling; and
\textbf{language organization or advocate,} an organization or person holding or working on language resources of various types, formats, and languages.
The published version of the catalogue submission form provides a variation on the main set of questions depending on the selected resource type. 

All entry submissions request information about the languages and locations of the resource as well as contact information for a representative, owner, or custodian of the resource.
Further questions are added for primary sources and processed datasets, including the availability of the resource and legal considerations for using the data, such as licenses and personally identifiable information (PII), the type of data it contains, and the medium of the data.


\subsubsection{General Information}
The form starts with the selection of the source type and updates the questions once a type is selected.
The first section requests general information such as the resource name, a unique identifier to use in searching the catalogue, and the resource's homepage for further information.
The form provides additional space for a description of the resource to display when searching the catalogue.

\subsubsection{Languages and Locations}
We designed the \textit{Languages and Locations} section to accommodate various degrees of granularity in order to support our goal of representativeness, evaluate the degree to which we achieve that goal, and maximize the usability of the catalogue beyond this current work.
Entry submitters may select any and all languages represented in the resource from prepared drop-down lists of the target BigScience language groups, with additional lists for the Indic and Niger-Congo language families to further specify individual languages, as well as any other languages as defined by the BCP-47 standard.\footnote{\url{https://tools.ietf.org/rfc/bcp/bcp47.txt}} 
The form also provides space for submitting comments about the language variety, such as the resource containing dialect data or code-switching. 
Similarly, information about the geographical origin of the data (as defined by the primary location of the language creators whose data is represented in the resource) may be answered using a drop-down list of macroareas ranging from worldwide to continents to regions such as Western Africa or Polynesia in addition to a list of specific countries, nations, regions, and territories.  

\subsubsection{Representative, Owner, or Custodian}
Responsible dataset creation includes respecting the rights of the person or organization that either owns or manages the data source, whom we refer to as the \textit{data custodian}.
The form gives the option to link the current resource being submitted to an existing organization in the catalogue via a drop-down list. 
If an existing organization entry is not linked, the remaining questions cover the name, type, location, and contact information of the data custodian.
This information supports our own and future catalogue users' efforts to understand local legal structures that may apply to the resource, communicate with data custodians about how their data is being used, and request permission for uses beyond those stated in published licenses. 

\subsubsection{Availability of the Resource}

For primary sources and existing datasets, the form requests information for how the data may be obtained.
The first question asks whether the data may be downloaded online with or without contacting the data custodian first.
Depending on the response, the form asks for either the URL to download the data or contact information for the data query.
In characterizing the licenses or terms of use for the data, the form first asks whether the resource is accompanied by an explicit license. 
If the license or terms are known, the submitter may select a description such as public domain, research use, non-commercial use, or do not distribute. 
Submitters can also select relevant licenses from a drop-down list of frequently used licenses or may copy the terms or license text into the form. 
If the licensing terms are unknown or unclear, the form requests that the submitter give their best assessment of whether the data can be used to train models while respecting the rights and wishes of the data creators and custodians.

In order to remove PII at the later stage of processing the data, we define three categories of PII, drawing from the standards laid out in the US Health Insurance Portability and Accountability Act of 1996 (HIPAA)\footnote{\url{https://www.hhs.gov/hipaa/index.html}} and the EU General Data Protection Regulation.\footnote{\url{https://gdpr.eu/}} While only a portion of the data collected in the catalogue may be in the same jurisdiction as these regulations, they provide a starting point for specific examples of information types that may lead to the identification of an individual person in any context.
\textbf{General PII} includes names, physical and email addresses, website accounts with names or handles, dates (birth, death, etc.), full-face photographs and comparable images, and biometric identifiers (fingerprints, voice, etc.).
\textbf{Numeric PII} includes identifying numbers such as contact information, vehicle and device identifiers and serial numbers, IP addresses, medical or health plan numbers, and any other uniquely identifying numbers.
\textbf{Sensitive PII} includes descriptions of racial or ethnic origin, political opinions, religious or philosophical beliefs, trade-union membership, genetic data, health-related data, and data concerning a person's sex life or sexual orientation.
The form asks submitters to determine whether the data is likely to contain any of the kinds of PII described above, and if so, to estimate the likelihood of that kind of PII appearing in the data from very likely to none.

If there is some likelihood of the data containing a category of PII, the submitter is asked to select the specific kinds of information that may appear in the data from a drop-down list of the examples given above.
We advise the submitter that the default answer should be that the resource does contain PII unless there is a very good reason to believe otherwise, in which case we ask the submitter to justify their response.
Considering common sources, we predicted that two likely justifications for the resource not containing PII would be that the data is either fictional or general knowledge that is not written by or referring to private persons.
We added these to the form as prepopulated answers, but the submitter may also give their own answer.

\subsubsection{Primary Source Type}
If the submission is a primary source, the form provides a section for describing the kind of data that the resource contains.
We provide options using drop-down lists for two kinds of resources:
\textbf{collections}, which may contain books or book publishers, scientific articles and journals, news articles, radio programs, movies and documentaries, or podcasts, and
\textbf{websites}, which may include social media, forums, news or magazine websites, wikis, blogs, or content repositories. 
The form provides functionality for characterizing other kinds of resources and giving additional examples of collections and websites using a free response input.

If the submission is a processed language dataset, the section appears in the form as \textit{Primary Sources of the Processed Dataset}. 
If the dataset contains original data, no further questions appear.
If the data is taken from one or more primary sources, the form presents questions about those sources, such as if the primary sources are available to investigate through documentation or description or being openly available.
The form provides a drop-down menu of primary sources already documented in the catalogue to link the processed dataset to if they are part of the dataset and another drop-down menu to describe the types of data in the primary sources. 
The final question concerns the licensing information of the primary sources, which may be different from the licensing information of the dataset itself.
We expect that most dataset licenses are compatible with their source material through either open licenses or prior consent, though there are also cases where it is unclear that the dataset respects the terms of the source material or even directly violates those terms.

\subsubsection{Media Type, Format, Size, and Processing}
The final section of the form focuses on the technical aspects of the resource.
The submitter may indicate the medium of the data, such as text, audiovisual data, or images, or a combination, and the technical details about the data format, such as the file type or distribution format.
If the data includes text, the form produces an additional question on whether the text was transcribed from another media format such as audiovisual data or images.
While most datasets appear with metadata about the size of the data in terms of megabytes or gigabytes, providing this kind of size estimate for primary sources is more difficult.
Instead, we asked submitters to provide a more descriptive estimate of the amount of data, starting with asking for the unit of data in terms of either articles, posts, dialogues, episodes, books, webpages, or some other description that submitters could provide themselves.
From this unit, the form asks for estimates of the number of instances in the resource and the number of words in the instance using ranges of magnitudes of 10.
After having filled out the form for a resource, submitters may review their answers as a JSON dictionary before saving the entry to the catalogue.

\section{Community Hackathons}


We organized hackathons for specific communities and regions of the world based on the availability of organizers and their familiarity with the communities, namely African languages (in collaboration with Masakhane)\footnote{\url{https://www.masakhane.io/}}, Asian languages (in collaboration with Machine Learning Tokyo)\footnote{\url{https://www.mlt.ai/}}, Basque, English in the Americas and Europe, English in the Indo-Pacific Region, and Spanish in Latin America (in collaboration with LatinX in AI)\footnote{\url{https://www.latinxinai.org/}}.
These hackathons took place online in October, November, and December of 2021, lasting one to six hours. A BigScience member would interact with participants and be available to answer questions that might arise while filling the form, and also to discuss particular resources or institutions.

The hackathons were announced using social media, in particular the BigScience Twitter account\footnote{\url{https://twitter.com/BigscienceW}} and also the accounts of the relevant partner organizations. 
Although the form requires a name and email in order to save an entry to the catalogue, we did not collect further demographic information during the hackathons in order to create the lowest barrier to entry for participation.
Instead, we sent out a short, 10-question survey to all participants after the hackathons. 


\section{Results}

\subsection{Hackathon Participation}

Across the hackathons, 41 participants submitted resources to the catalogue, of which 11 responded to the survey we sent out following the final hackathon. 
The first questions focused on participants' professional context, such as the country they are located in, their field of study, and their current stage in their career. The responses showed diversity in both geographical location and career stages.
Four respondents were located in Spain, with 3 specifically in the Basque Country, while the rest were spread across France, Japan, Kenya, Singapore, Sweden, Taiwan, and the USA. 
Respondents' career stage ranged from undergraduate student to a senior level position in industry, though most (7) listed an academic position. 
The most common research interests included NLP (8), data science (5), and linguistics (4).
Other interests included library and/or information science, ethics/safety, recommendation systems, vision, creative AI, and optimization and compression techniques.

The remaining questions concerned participants' experiences before and during the hackathons.
Most participants heard of the hackathons through either BigScience internal channels or through the communities and organizations that collaborate with BigScience.
Only two respondents listed social media as their entry point for the hackathons.
Most respondents (6) only added resources for languages that they were native or advanced speakers of.
Three respondents contributed resources that covered almost all of the BigScience languages, most of which they had no familiarity with.
In describing their motivations for participating in the hackathons, the most common reasons included developing the BigScience dataset, supporting under-resourced languages in general, and improving the coverage of a particular language.

\subsection{Gathered Resources}
\begin{figure}[t]
    \includegraphics[scale=.47]{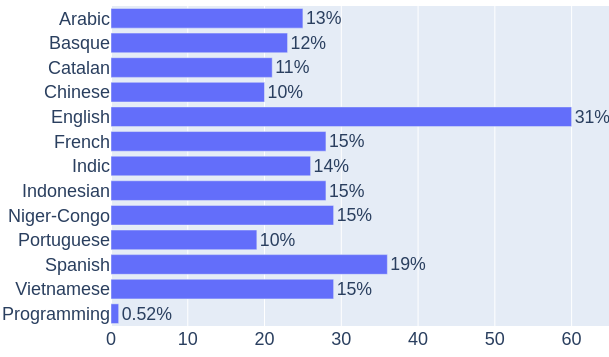}
    \caption{Relative distribution of the entries over the BigScience target languages.}
    \label{fig:bar_chart}
\end{figure}

As per 14 December 2021, the catalogue contains 192 entries with 432 different language tags (each entry can have multiple language tags). The most frequent language tags are those of the BigScience target language groups. The distribution of the target language groups across the entries in the catalogue is shown in Figure~\ref{fig:bar_chart} (note that due to multilingual resources, the percentages do not add up to 100\%). English is the most frequent language across all entries. The most frequent varieties of Arabic are Modern Standard Arabic and Classical Arabic (13 entries and 5 entries; all the other varieties have 2 or less entries), the most frequent Indic languages are Hindi, Bengali, Telugu, Tamil and Urdu (15, 11, 9, 9 and 8 entries; the other languages have between 4 and 8 entries), and the most frequent Niger-Congo languages are Swahili, Igbo, Yoruba and isiZulu (9, 7, 6 and 4 entries; the other languages have 3 or less entries).

On the other hand, 380 languages were tagged only in 1 or 2 entries. 
However, some of these languages actually belong to a broader target language group; they include 10 languages from the Niger-Congo group (Sesotho, Kirundi, Bambara, Kinyarwanda, Chi Chewa, Wolof, Twi, Lingala, ChiShona, Kikuyu) and 12 varieties of Arabic (Algeria, Djibouti, Gulf, Egypt, Levant, Libya, Mauritania, Morocco, North Africa, Somalia, South Sudan, Sudan). So excluding these languages, 358 languages were tagged only once or twice.  

The languages in the catalogue show a clear bias towards certain languages. Taken together, English and Spanish account for about half of the target languages recorded in the catalogue. On the other hand, Chinese is included in fewer entries than languages that are less widely spoken (e.g. French, Spanish and Vietnamese; see \cite{ethnologue2021}). This imbalance is the result not only of the varying availability of sources across different languages, but also of the countries of origin and linguistic expertise of the BigScience contributors and hackathon participants.

While we allowed for users to provide different levels of granularity in defining the geographic location of each source, we did not have a minimum requirement for the location response. The form saves all responses in terms of macroscopic area (e.g. a continent or a macroregion within a continent), country, region within a country or some combination of all the options above to the same list in the catalogue.
As a results of this design decision, we cannot systematically report on all geographical categories.
For this reason, when analysing the geographical distribution of the recorded languages, we report on the first geographical area provided for each entry. This was usually a macroscopic area, but in some cases only a specific country or region was provided. These geographical locations have then been manually assigned to their respective macroscopic area. The resulting distribution is shown in Table~\ref{tab:distribution_languages_geo}. More than half of all the primary language locations of the entries belong to Asia and Europe.

We further analyzed the locations of English, French, Spanish and Portuguese entries to investigate how different regional varieties are represented. All the locations (not just the first ones) were manually grouped into macroscopic areas and are shown in Table~\ref{tab:distribution_EnFrEsPr}. We can see that all these languages are well represented in their European varieties. However, each language also has a good amount of entries from other geographical areas (which are specific of each language) as well as several entries that were tagged as `World-wide' (resources that include examples of the target language from multiple locations).
\begin{table}[t]
    \centering\small
    \begin{tabular}{l c c}
         \toprule
         \multirow{2}{*}{Language location} & \multirow{2}{*}{\#} & Percentage \\
          & & of all entries\\
         \midrule
         Africa & 18 & 9.38\%\\
         Americas* & 3 & 1.56\%\\
         Asia & 61 & 31.77\%\\
         Europe & 46 & 23.96\%\\
         Latin America and the Caribbean & 17 & 8.85\%\\
         Middle East and North Africa & 4 & 2.08\%\\
         North Africa & 2 & 1.04\%\\
         North America & 11 & 5.73\%\\
         Oceania & 5 & 2.60\%\\
         World-wide & 21 & 10.94\%\\
         \bottomrule
    \end{tabular}
    \newline\raggedright\footnotesize
    $^*$ entries not specifying if N. Am. or Lat. Am. and the Car.
    \caption{Distribution of data creators' locations over geographic regions (only first location for each entry).}
    \label{tab:distribution_languages_geo}
\end{table}
\begin{table}[ht]
    \centering\small
    \begin{tabular}{l c c c c}
         \toprule
         \multirow{2.5}{*}{Location} & \multicolumn{4}{c}{Languages}\\
         \cmidrule(lr){2-5}
          & En. & Fr. & Sp. & Port. \\
          \midrule
          Africa* & 6 & 4 & 0 & 1 \\
          Americas$^\dag$ & 0 & 1 & 2 & 1 \\
          Asia & 10 & 0 & 0 & 1 \\
          Europe & 13 & 13 & 11 & 5 \\
          Latin America and the Carib. & 3 & 0 & 15 & 2 \\
          North America & 13 & 1 & 2 & 1 \\
          Oceania & 5 & 0 & 0 & 0 \\
          World-wide & 16 & 11 & 10 & 11 \\
         \bottomrule
    \end{tabular}
    \raggedright\footnotesize
    $^*$ including entries from North Africa; no entries from Middle East were recorded for these languages
    
    $^\dag$ entries not specifying if N. Am. or Lat. Am. and the Car.\\
    \caption{Distribution of entries in English, French, Spanish and Portuguese across continents.}
    \centering\small
    \label{tab:distribution_EnFrEsPr}
\end{table}
In terms of source types, the largest share of entries were primary sources. Of the 192 catalogue entries, 98 (51\%) are primary sources, 64 (33\%) are processed sources and 30 (16\%) are organizations. Table~\ref{tab:distribution_types} shows the distribution of the source types across the target language groups. With the exception of Catalan, Indic and Vietnamese, the target language groups have more primary sources than secondary sources.
\begin{table}[ht]
    \centering\small
    \begin{tabular}{l c c c}
         \toprule
         \multirow{2}{*}{Languages} & \multicolumn{3}{c}{Types}\\
         \cmidrule(lr){2-4}
          & Primary & Processed & Org. \\
          \midrule
          Arabic & 13 & 3 & 9 \\
          Basque & 15 & 0 & 8 \\
          Catalan & 1 & 14 & 6 \\
          Chinese & 9 & 4 & 7 \\
          English & 29 & 13 & 18 \\
          French & 13 & 4 & 11 \\
          Indic & 8 & 11 & 7 \\
          Indonesian & 15 & 8 & 5 \\
          Niger-Congo & 11 & 5 & 13 \\
          Portuguese & 7 & 3 & 9 \\
          Programming & 1 & 0 & 0 \\
          Spanish & 17 & 2 & 17 \\
          Vietnamese & 8 & 15 & 6 \\
         \bottomrule
    \end{tabular}
    \caption{Distribution of the target languages in the catalogue across source types.}
    \label{tab:distribution_types}
\end{table}

The largest share of sources recorded are stewarded by non-commercial entities. Table~\ref{tab:distribution_custodians} shows the distribution of entries across custodian types. University and research institutions are the most frequent custodian type (23.44\%), followed by commercial entities (21.35\%) and nonprofit entities/NGOs (13.5\%). In 24 records (12.5\%) the custodian is missing. 
\begin{table}[ht]
    \centering\small
    \begin{tabular}{l c}
         \toprule
         \multirow{1}{*}{Custodian type} & \multirow{1}{*}{\#}\\
         \midrule
        University or research institution & 45 \\
        Commercial entity & 41 \\
        Nonprofit / NGO & 26 \\
        Private individual & 20 \\
        Government organization & 17 \\
        Library, museum or archival institute & 16\\
        Community (incl. online) & 2 \\
        Startup & 1 \\
         \bottomrule
    \end{tabular}
    \caption{Distribution of custodian types.}
    \label{tab:distribution_custodians}
\end{table}
In terms of geographic diversity of the custodians, the catalogue records 164 different locations for custodians (28 entries do not have a custodian location). The custodian locations reflect the diversity of the catalogue, but also show that a large share of the resources are hosted in the US and in a few European countries. The top 12 custodian locations are shown in Table~\ref{tab:distribution_custodians_t12}.
All the other locations were recorded only twice (Argentina, Bangladesh, Brazil, Japan, Jordan, Mozambique, Nepal, Nigeria, Taiwan) or once (Bolivia, Burundi, Czech Republic, Ethiopia, Hong Kong, Ireland, Italy, Kenya, Luxembourg, Mexico, Netherlands, Peru, Saudi Arabia, Scotland, Thailand, Turkey, United Arab Emirates).

\begin{table}[ht]
    \centering\small
    \begin{tabular}{l c|l c}
         \toprule
         \multirow{1}{*}{Custodian location} & \multirow{1}{*}{\#} & \multirow{1}{*}{Custodian Location} & \multirow{1}{*}{\#}\\
         \midrule
         Spain & 27 & France & 9 \\
         USA & 22 & South Africa & 6 \\
         Vietnam & 14 & UK & 5 \\
         Indonesia & 14 & Australia & 4 \\
         India & 11 & Germany & 4 \\
         Colombia & 10 &  China & 3 \\
         \bottomrule
    \end{tabular}
    \caption{Top 12 most frequent custodian locations.}
    \label{tab:distribution_custodians_t12}
\end{table}

As regards licensing, the available metadata suggests that the hackathon participants made an effort to collect sources with an open license or without copyright. Table~\ref{tab:distribution_licenses} shows the frequency of license properties in the catalogue (note entries may have multiple license properties). Sources with an open license or in the public domain account for some 37\% of the sources in the catalogue. 
However, for another 37\% of the entries in the catalogue no licensing properties were recorded.


\begin{table}[ht]
    \centering\small
    \begin{tabular}{l c c}
         \toprule
         Licensing & \multirow{2}{*}{\#} & Percentage \\
         properties & & of all entries\\
         \midrule
         Missing & 71 & 37\% \\
         Open license & 56 & 29\%\\
         Copyright & 30 & 16\%\\
         Non-commercial use & 18 & 9\%\\
         Public domain & 18 & 9\%\\
         Research use & 10 & 5\%\\
         Multiple licenses & 7 & 4\%\\
         Do not distribute & 2 & 1\%\\
         \bottomrule
    \end{tabular}
    \caption{Distribution of licensing properties.}
    \label{tab:distribution_licenses}
\end{table}
Finally, the results of our PII metadata analysis highlights the importance of proper handling of sensitive personal information. More than half of the catalogue has PII of some sort, as shown in Table~\ref{tab:distribution_pii}. Another 33\% of the catalogue has either unclear information or no metadata about PII, which calls for a cautious approach with regard to PII management. Only 13\% of the catalogue has no PII, according to the metadata recorded during the hackathons. 
\begin{table}[ht]
    \centering\small
    \begin{tabular}{l c c}
         \toprule
         \multirow{2}{*}{Contains PII} & \multirow{2}{*}{\#} & Percentage \\
         & & of all entries\\
         \midrule
         Yes & 84 & 44\% \\
         Unclear & 48 & 18\%\\
         Answer Missing & 30 & 16\%\\
         No & 25 & 13\%\\
         Yes (text author's name only) & 18 & 9\%\\
         \bottomrule
    \end{tabular}
    \caption{Distribution of entries with PII or sensitive information.}
    \label{tab:distribution_pii}
\end{table}


\section{Discussion}

As a result of our efforts, we were successfully able to create an openly available catalogue of 192 data sources with at least 10\% of the entries representing each of our target languages (with the exception of programming languages) in locations around the world.
The bulk of these resources are primary sources, presenting opportunities to collect data in these languages in new contexts and topics.
Together the resource custodians themselves cover a wide range of geographic and institutional contexts.
Our participants' initial efforts in estimating the presence of PII in the resources suggest that PII is significant across the sources and should be an important consideration in technologies built on these data sources. 
Furthermore, the documentation collected on these resources will continue to be accessible for the BigScience project as well as other projects that use the resources.

\subsection{Challenges in creating the catalogue}

We encountered a number of challenges while creating the catalogue. The hackathons were successful in drawing a large range of entries, but our analysis shows that the catalogue entries are still largely concentrated on highly resourced languages and locations with a long tail of single instances of representation.

\paragraph{Recruiting volunteers.} 
The motivations for volunteer participation in projects like our catalogue have been explored in citizen science and crowdsourcing research across disciplines such as astronomy \cite{raddick2013galaxy}, biology \cite{berger2017wildbook} and history research \cite{causer2014crowdsourcing}.
A common finding of many such studies and supported in recent NLP projects such as EleutherAI's Evaluation Harness \cite{eval-harness} and Google's Big Bench \cite{bigbench} is that a small number of contributors contribute the majority of data and a large number of contributors only contribute once \cite{segal2016intervention}. 
These observations emphasize the importance of having a large base number of contributors for events such as our hackathons, but our total number of 41 participants fell short of desired numbers given the scope of our project.
Despite advertising the hackathons publicly, the results of our participant survey suggest that most of the participants were already part of the BigScience initiative or partner organizations. Future hackathon efforts should focus on outreach through partner organizations and should allow for more time for news of the events to travel. The actual (or perceived) difficulty of contributing to the catalogue may further bar participation from individuals not affiliated with BigScience. Additionally, the motivation for contributing to this kind of activity may also suffer as a result of a broader undervaluing of data-related work in NLP and ML \cite{sambasivan_etal2021data_cascades}.

\paragraph{Creating catalogue entries.} 
In the participants survey, we asked respondents about the challenges they faced while contributing to the catalogue.
They noted difficulties in finding specific metadata, in finding appropriate resources, and with the catalogue infrastructure.
Respondents had trouble determining which resources were appropriate for the catalogue given possible conflicts in terms of later use in training models and finding sources for particular languages. 
Even if respondents did have a resource to submit, they had difficulty finding or estimating pieces of metadata, particularly information about the custodian and the amount of data as well as the license, type and format of the data, and the original curation rationales.
Primary data sources in particular often lack metadata about PII and licensing, as reflected in \S5.2.
This challenge has been identified by libraries and archives as creating metadata to describe collections is one of their core missions. 
However, \newcite{padilla_thomas_2019_3152935} found that there is often a gap between the detail of metadata at the item level and metadata for collections made available ``as data'', suggesting that this challenge may not be easily addressed with existing infrastructure.
Respondents also wanted improved features for the catalogue itself, such as fuzzy-search to find existing entries and visualization for the relations between the resources. 
For future hackathons, respondents suggested language-specific channels for sharing resources and information, more accessible times for the events, and support for uploading CSV files for highly multilingual resources.

\subsection{Limitations of the catalogue}


We identified limitations in our catalogue with regards to the language coverage, the scope of metadata, and the resource management.
Our catalogue only covers a small fraction of world languages. Missing languages include some of the most highly spoken languages as well as the majority of under-represented languages. In addition, the distribution of the BigScience target languages in the catalogue is not uniform. Nonetheless, the collaborative effort of the hackathon has lead to a good degree of diversity among the language varieties covered, especially with regard to English, French, Spanish and Portuguese.

The metadata in our catalog does not include more in-depth information regarding some characteristics of the data. For example, there is no explicit information that informs the quality of the dataset itself, although asking crowdworkers to additionally provide dataset quality information is arguably challenging and time consuming. Similarly, the language characteristics such as style, dialect or geographical location is not well captured in our metadata. The challenge with recording the dialects and the geographical location of the data is that the sources may include examples from a variety of combinations of different dialects and/or regions, which makes it difficult to create a standardised classification system that can be applied to every source. Furthermore, information about the geographical location of the languages in the sources may not be easily accessible or available. A loosely structured ontology for recording dialects and geographical locations in the catalogue provides users with the flexibility to record the metadata of each specific source. The downside of this approach is that it becomes more difficult to extract precise information on the distribution of the dialects and geographical locations from the catalogue.



\subsection{Recommendations}

In reflecting on our experience in creating the catalogue, we suggest recommendations for future work regarding designing tasks for crowdsourcing efforts, engaging the broader data ecosystem, and using the catalogue. 
The task of completing a catalogue entry proved to be reasonably complex, requiring both domain knowledge of potential sources (or knowledge on how to identify these) and some understanding of how to identify the metadata needed for the catalogue entry. Future initiatives may explore breaking down these tasks for creating or reviewing catalogue entries into smaller parts. A potential task for volunteers could include recording or correcting that information related to language, PII, and licensing were shown to be inconsistent or unanswered in the catalogue. Based on the recommendations of crowdsourcing task designers in the cultural heritage sector, it may help to also develop different roles for volunteers, such as a reviewer role \cite{Ridge20218}.


Establishing collaborations with data custodians, especially libraries and archives, who have existing roles in curating and describing collections could result in easier access to metadata and data resources provided by these institutions and also support the development of ethical best practices\cite{jo_gebru2020archives} and metadata standardization.
A standardized machine-readable metadata schema would allow for more accessible aggregation across different records, though selection and adoption of one standard will take time and organization from many stakeholders.
One such example, DataCite\footnote{\url{https://schema.datacite.org/}} provides a core metadata schema that has been adopted across many data and software repositories and allows for easy comparison.

Future users of the catalogue and the data referenced within should consider the limitations described in \S6.2. Whilst the catalogue is open, the underlying data in the catalogue have their own licensing and usage restrictions that future users must abide by, e.g. whether the license precludes commercial use of data. 
Appropriate handling of PII in the data sources should also be included in any future plans for the catalogue, with care taken for the detection and implications of the different types of PII outlined in \S3.1.4.




\section{Conclusion}

We have presented our design processes, our human-centered metadata collection efforts, and our resulting successes and challenges in creating the BigScience catalogue targeting 13 language groups.
Next steps for the catalogue include filling in the missing information for the current entries and adding more entries to continue working toward our goal of greater representation across languages and regions.
We will continue to maintain the catalogue while developing the BigScience dataset, so that others may use it as a reference for future dataset and modeling endeavors.
We hope to encourage others to follow conscientious documentation practices prior to releasing data collections, especially in large-scale settings.

\section{Acknowledgements}

We are grateful to all of the participants of the hackathons, without whom this work would not have been possible, and to all of our BigScience colleagues who have provided valuable feedback on this work.

\begin{figure*}[t]
    \centering
    \includegraphics[scale=.7]{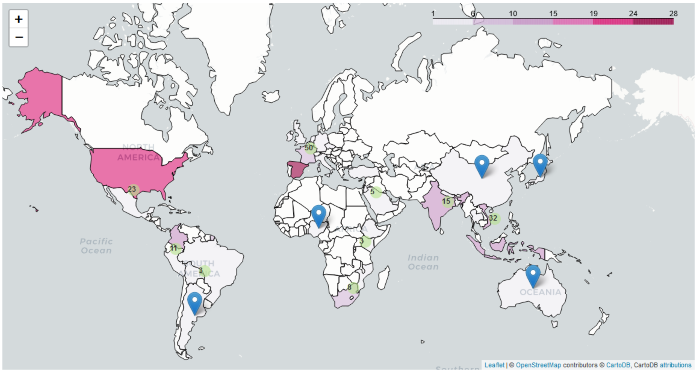}
    \caption{Geographical visualization of the locations of entries' data custodians.}
    \label{fig:map}
\end{figure*}

\section*{Appendix: Additional Features of the Catalogue}

The form also provides a mode for a second participant to review and validate entries already submitted to the catalogue. 
After selecting an entry, the form updates with the responses originally submitted for that entry with a validation checkbox at the end of each section.
The validator may review and edit the selections for each question and mark the section as validated. 
Once each section has been reviewed, the validator may save their work.
Already validated entries will include a note indicating that the entry has been validated and allow the validator to review either the original entry or later entries listed by their save date. 

A third mode allows participants to visualize and review the entries in the catalogue using filters.
An interactive map of the world shows the number of entries submitted by various geographical levels of detail, such as region or country, for either the location of the data creators or the location of the data custodians.
As shown in the snapshot of the map in Figure~\ref{fig:map}, the color gradient indicates the number of entries by country and the location markers indicate regions that can be zoomed in on for more details.
Both the map and a pie chart showing the proportion of entries by language may be filtered using one of the many properties produced by the form such as the resource type, the license type, or the media type.
Each entry returned by the filter may be selected to review its description as provided in the first section of the form.

\section{Bibliographical References}\label{reference}

\bibliographystyle{lrec2022-bib}
\bibliography{lrec2022}

\section{Language Resource References}
\label{lr:ref}
\bibliographystylelanguageresource{lrec2022-bib}
\bibliographylanguageresource{languageresource}

\end{document}